\documentclass[runningheads]{llncs}
\usepackage{graphicx}
\usepackage{comment}
\usepackage{amsmath,amssymb} 
\usepackage{color}
\usepackage{stmaryrd}
\usepackage{dsfont}
\usepackage{multirow}
\usepackage{booktabs}
\usepackage{caption}
\usepackage{breakcites}
\usepackage{wrapfig}
\newcommand{\tabincell}[2]{\begin{tabular}{@{}#1@{}}#2\end{tabular}}  
\usepackage{siunitx}
\usepackage{caption}
\usepackage{booktabs}
\usepackage{array}
\usepackage{bm}
\makeatletter
\newcommand{\printfnsymbol}[1]{%
  \textsuperscript{\@fnsymbol{#1}}%
}
\makeatother

\begin{document}
\pagestyle{headings}
\mainmatter
\def\ECCVSubNumber{233}  

\title{Temporal Distinct Representation Learning \\ for Action Recognition} 


\titlerunning{Temporal Distinct Representation Learning}
\author{Junwu Weng\thanks{Equal contribution. This work is done when Junwu Weng is an intern at Youtu Lab.} \inst{1,3} \and
Donghao Luo\printfnsymbol{1}\inst{2} \and 
Yabiao Wang\inst{2} \and
Ying Tai\inst{2} \and
Chengjie Wang\inst{2} \and
Jilin Li\inst{2} \and
Feiyue Huang\inst{2} \and 
Xudong Jiang\inst{3} \and
Junsong Yuan\inst{4}
}  
\authorrunning{J. Weng, D. Luo, et al.}
%
\institute{Tencent AI Lab \and Tencent Youtu Lab \and
School of EEE, Nanyang Technological University \and
Department of CSE, The State University of New York, Buffalo \\ \{calweng, michaelluo, caseywang, yingtai, jasoncjwang, \\ jerolinli, garyhuang\}@tencent.com \\ 
exdjiang@ntu.edu.sg~~~~jsyuan@buffalo.edu}

\maketitle

\begin{abstract}


Motivated by the previous success of Two-Dimensional Convolutional Neural Network (2D CNN) on image recognition, researchers endeavor to leverage it to characterize videos. However, one limitation of applying 2D CNN to analyze videos is that different frames of a video share the same 2D CNN kernels, which may result in repeated and redundant information utilization, especially in the spatial semantics extraction process, hence neglecting the critical variations among frames. In this paper, we attempt to tackle this issue through two ways. 1) Design a sequential channel filtering mechanism, i.e., Progressive Enhancement Module (PEM), to excite the discriminative channels of features from different frames step by step, and thus avoid repeated information extraction. 2) Create a Temporal Diversity Loss (TD Loss) to force the kernels to concentrate on and capture the variations among frames rather than the image regions with similar appearance. Our method is evaluated on benchmark temporal reasoning datasets Something-Something V1 and V2, and it achieves visible improvements over the best competitor by $2.4\%$ and $1.3\%$, respectively. Besides, performance improvements over the 2D-CNN-based state-of-the-arts on the large-scale dataset Kinetics are also witnessed.

\end{abstract}

\section{Introduction}
\label{sec:intro}

Owing to the computer vision applications in many areas like intelligent surveillance and behavior analysis, how to characterize and understand videos becomes an intriguing topic in the computer vision community. To date, a large number of deep learning models~\cite{c3d,tsn,tsm,trn,nonlocal,convlstm,lu2018deep,lu2019see,lu2020learning} have been proposed to analyze videos. The RNN-based models~\cite{convlstm,valstm} are common tools for sequence modeling for its sequential nature of visual representation processing, by which the order of a sequence can be realized. However, in these models the spatial appearance and temporal information are learned separately. Motivated by the success in image recognition, Convolutional Neural Network (CNN) becomes popular for video analysis. 3D CNNs~\cite{nonlocal,i3d,slowfast,c3d} are widely used in video analysis as they can jointly learn spatial and temporal features from videos. However, their large computational complexities impede them from being applied in real scenarios. In contrast, 2D CNNs are light-weight, but do not bear the ability for temporal modeling. To bridge the gap between image recognition and video recognition, considerable 2D-CNN-based researches~\cite{tsn,tbn,tsm,trn,abm} recently attempt to equip the conventional 2D CNNs with a temporal modeling ability, and some improvements are witnessed.
 
However, another direction seems to be less explored for 2D-CNN-based video analysis, namely diversifying visual representations among video frames. Although the 2D CNN takes multiple frames of a video at once as input, the frames captured from the same scene share the same convolution kernels. A fact about CNN is that each feature channel generated by the kernel convolution from the high-level layers highly reacts to a specific semantic pattern. Hence, with 2D CNN, the yielded features from different frames may share multiple similar channels, which thereafter results in repeated and redundant information extraction for video analysis. If the majority part of frames is background, these repeated redundant channels tend to describe the background scene rather than the regions of interest. This tendency may lead to the ignorance of the motion information which can be more critical than the scene information for action understanding~\cite{jiang2012trajectory,wang2016trans,wang2018context,choi2019mall}. Besides, the customary strategy that features from different frames of a video are learned under the same label of supervision will make this issue even more severe. We observe that for one temporal reasoning dataset like Something-Something~\cite{goyal2017something}, video samples under the same category are from various scenes and the actions therein are performed with various objects. The scene and object information may not be directly useful for the recognition task. Thus, a 2D-CNN-based method like TSN~\cite{tsn} is easy to overfit as the model learns many scene features and meanwhile neglects the variations among frames, {\it e.g.} the motion information. We state that due to this redundant information extraction, the previously proposed temporal modeling method cannot fully play its role. In this paper, we propose two ways to tackle the issue.
 
We first introduce an information filtering module, i.e., Progressive Enhancement Module (PEM), to adaptively and sequentially enhance the discriminative channels and meanwhile suppress the repeated ones of each frame's feature with the help of motion historical information. Specifically, the PEM progressively determines the enhancements for the current frame's feature maps based on the motion observation in previous time steps. This sequential way of enhancement learning explicitly takes the temporal order of frames into consideration, which enables the network itself to effectively avoid gathering similar channels and fully utilize the information from different temporal frames. After PEM, we set a temporal modeling module that temporally fuses the enhanced features to help the discriminative information from different frames interact with each other. 

Furthermore, the convolution kernels are calibrated by the Temporal Diversity Loss (TD Loss) so that they are forced to concentrate on and capture the variations among frames. We locate a loss right after the temporal modeling module. By minimizing the pair-wise cosine similarity of the same channels between different frames, the kernels can be well adjusted to diversify the representations across frames. As the TD Loss acts as a regularization enforced to the network training, it does not add an extra complexity to the model and keeps a decent accuracy-speed tradeoff. 
 
We evaluate our method on three benchmark datasets. The proposed model outperforms the best state-of-the-arts by $2.4\%$, $1.3\%$ and $0.8\%$ under the $8f$ setting on the Something-SomethingV1, V2 and the Kinetics400 datasets, respectively, as shown in Table~\ref{tbl:sth} and Table~\ref{tbl:k400}. The proposed PEM and TD Loss outperform the baseline by $2.6\%$ and $2.3\%$ on Something-Something V1, respectively. The experimental results demonstrate the effectiveness of our proposed 2D-CNN-based model on video analysis. 
\\
\\
Our contributions can be summarized as follows:
 
 \begin{itemize}\itemsep0pt
 	\item[$\bullet$] We propose a Progressive Enhancement Module for channel-level information filtering, which effectively excites the discriminative channels of different frames and meanwhile avoids repeated information extraction.
 	\\
	\item[$\bullet$] We propose a Temporal Diversity Loss to train the network. The loss calibrates the convolution kernels so that the network can concentrate on and capture the variations among frames. The loss also improves the recognition accuracy without adding an extra network complexity.
\end{itemize}

\section{Related Work}
\label{sec:works}

  \subsubsection{2D-CNNs for Video Analysis} Due to the previous great success in classifying images of objects, scenes, and complex events~\cite{imagenet,verydeep,goingdeep,resnet}, convolutional neural networks have been introduced to solve the problem of video understanding. Using two-dimensional convolutional network is a straightforward way to characterize videos. In Temporal Segment Network~\cite{tsn}, 2D CNN is utilized to individually extract a visual representation for each sampled frame of a video, and an average pooling aggregation scheme is applied for temporal modeling. To further tackle the temporal reasoning of 2D CNNs, Zhou {\it et.al.} proposed a Temporal Relational Network~\cite{trn} to hierarchically construct the temporal relationship among video frames. Ji {\it et.al.} introduced a simple but effective shift operation between frames into 2D CNN, and proposed the Temporal Shift Module~\cite{tsm}. Following the same direction, the Temporal Enhancement Interaction Network (TEINet)~\cite{tei} introduces a depth-wise temporal convolution for light-weight temporal modeling. Similar methods include Temporal Bilinear Network~\cite{tbn} and Approximate Bilinear Module~\cite{abm}, which re-design the bilinear pooling for temporal modeling. These methods attempt to equip the 2D CNN with an ability of temporal modeling. However, one neglected limitation of the 2D CNN is the redundant feature extraction among frames or the lack of temporal representation diversity. This is the battle field to which our proposed method is engaged. We first propose a Progressive Enhancement Module before temporal modeling to enhance the discriminative channels and meanwhile suppress the redundant channels of different frames sequentially. Furthermore, after the temporal modeling module, we create a Temporal Diversity loss to force the convolution kernels to capture the variations among frames.
 
 \subsubsection{Channel Enhancement} The idea of enhancing the discriminative channels for recognition first appears in image recognition. In Squeeze-and-Excitation Network (SENet)~\cite{hu2018squeeze}, an attention sub-branch in the convolutional layer is involved to excite the discriminative channels of frame's features. Inheriting from the SENet, to emphasize the motion cues in videos, TEINet uses the difference between feature maps of two consecutive frames for channel-level enhancement learning. In our method, we expand the receptive field of this channel enhancement module. At each time step, the enhancement module is able to be aware of the motion conducted in previous frames, therefore avoiding activating the channels emphasized previously.
 
 \subsubsection{Diversity Regularization} In fine-grained image recognition, to adaptively localize discriminative parts, attention models are widely used. However, the previously proposed attention models perform poorly in classifying fine-grained objects as the learned attentions tend to be similar to each other. In~\cite{zheng2017learning,zhao2017diversified}, attention maps are regularized to be diverse in the spatial domain to capture the discriminative parts. In this paper, we take the temporal diversity of the feature maps into consideration and propose the Temporal Diversity Loss. The TD Loss directly sets the regularization on the visual representation of each frame to obtain the discriminative and dynamic features for video analysis.

\section{Proposed Method}
\label{sec:method}

\begin{figure}[t]
      \begin{center}
         \includegraphics[width=1.0\linewidth]{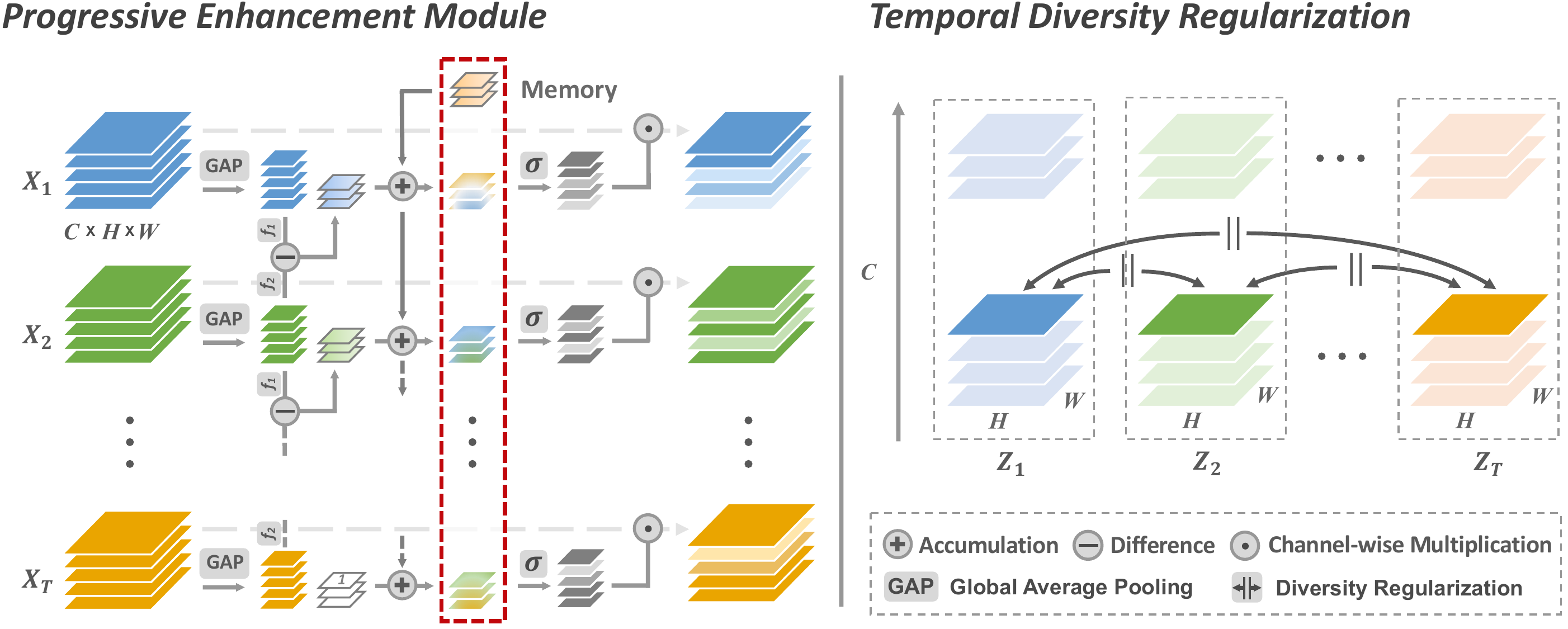}
      \end{center}
         \caption{An illustration of the proposed framework. In PEM, 1) features of each frame are GAP-ed and down-sampled to get a vector. 2) The differencing operation is performed on the vectors of each two consecutive frames. 3) The {\it memory} vector (in the red box) accumulates historical difference information. 4) With the Sigmoid function, the channel-level enhancement is obtained to excite discriminative channels of each frame. To compress the model complexity, the $1\times1$ convolution operation in $f(\cdot)$ reduces the vector dimensionality and the one before $\sigma(\cdot)$ recovers it back to $C$. In TD regularization, the same channels of each frame pair are regularized to be distinguished from each other.}
      \label{fig:framework}
   \end{figure}

In this section, we elaborate on the two contributions of this work. We first give the framework of the proposed method in Sec.~\ref{sec:framework}. The Progressive Enhancement Module is introduced in Sec.~\ref{sec:progressatt}. In Sec.~\ref{sec:tdloss}, the Temporal Diversity Loss for diverse representation modeling is described. The illustration of the whole framework is shown in Fig.~\ref{fig:framework}.

\subsection{Framework}
\label{sec:framework}

In our model, each video frame is represented by a tensor $\bm{X}^b_t \in \mathbb{R}^{C_b \times W_b \times H_b}$, which bears $C_b$ stacked feature maps with width $W_b$ and height $H_b$, and $b$ indicates the block index. In the following, we use $C$, $W$ and $H$ instead to simplify the notations. Given a video sequence/clip with $T$ sampled frames $V=\{\bm{X}_t\}^{T}_{t=1}$, the goal of our established deep network is to extract a discriminative visual representation of $V$ and predict its class label $k \in \{1, 2, ..., K \}$, where $K$ is the number of categories. Each block of the network takes the $T$ frames as input and outputs the feature maps for the next block, which is formulated as follows:

	\begin{equation}
         (\bm{X}^b_1, \dotsm, \bm{X}^b_T) = \mathcal{F}(\bm{X}^{b-1}_1, \dotsm, \bm{X}^{b-1}_T; \bm{\theta}^b), \label{eq:formulation}
     \end{equation}

\noindent where $\mathcal{F}$ is the feature mapping function, which involves the Progressive Attention for information filtering (Sec.~\ref{sec:progressatt}) and temporal information interaction. $\bm{\theta}$ is the block parameters to be optimized. The input and output of the network are denoted as $\bm{X}^0$ and $\bm{X}^B$, and $B$ is the total number of blocks.

The output feature maps $\{\bm{X}^B_t\}^{T}_{t=1}$ are gathered by average pooling, and are further fed into the Softmax function for category prediction. This mapping is defined as $\bm{\hat{y}} = \mathcal{G}(\bm{X}^B_1, \dotsm, \bm{X}^B_T)$, where $\bm{\hat{y}} \in [0,1]^K$ contains the prediction scores of $K$ categories. Therefore, the loss function is defined as

	\begin{equation}
 		\mathcal{L} = \mathcal{L}_c + \lambda\mathcal{L}_r = -\sum^{K}_{i=1} y_i \cdot log\,\hat{y}_i + \lambda\mathcal{L}_r, \label{eq:loss}
    \end{equation}
    
\noindent where $\mathcal{L}_c$ is the Cross Entropy Loss for category supervision. $y_i$ is the groundtruth label concerning class $i$, and it is an element of the one-hot vector $\bm{y} \in \{0,1\}^K$. $\mathcal{L}_r$ is the regularization term for network training, and $\lambda$ balances the importance between category supervision and network regularization. To enhance the temporal diversity of feature maps from different frames and thereafter to model the crucial motion information, the regularization term is defined as the Temporal Diversity Loss as depicted in Sec.~\ref{sec:tdloss}.

\subsection{Progressive Enhancement Module}
\label{sec:progressatt}

As discussed in Sec.~\ref{sec:intro}, one drawback of using 2D CNN for video analysis is that most of the kernels in one convolutional network are inclined to focus on repeated information, like scenes, across the features from different time steps, which cannot easily take full advantage of information from the video. The Progressive Enhancement Module (PEM) can sequentially determine which channels of each frame's features to focus on, and therefore effectively extract action related information. In each block, the feature maps $\{\bm{X}^{b-1}_t\}^{T}_{t=1}$ from the preceding block are first fed into the Progressive Enhancement Module for information filtering, as illustrated in Fig.~\ref{fig:framework}. Let $\bm{a}^b_t \in \mathbb{R}^{C}$ denote the enhancement vector to excite the discriminative channels of each frame. This operation is defined as

	\begin{equation}
		\bm{U}^b_t = \bm{X}^{b-1}_t \odot \bm{a}^{b}_t \,, \label{eq:excite}
	\end{equation}

\noindent where $\bm{U}^b_t$ is the $t$-th frame output of PEM in the $b$-th block, and $\odot$ is a channel-wise multiplication. For notational simplicity, we remove the block-index notation $b$ in the following description. 

The input feature maps $\{\bm{X}^{b-1}_t\}^{T}_{t=1}$ are first aggregated across the spatial dimensions by using Global Average Pooling (GAP), and the channel-wise statistics $\{\bm{x}_t\}^{T}_{t=1}$, $\bm{x} \in \mathbb{R}^C$ are then obtained. Each pair of neighboring frames in $\{\bm{x}_t\}^{T}_{t=1}$ is then fed into two individual $1 \times 1$ convolution operations $f_1$ and $f_2$ with ReLU activation, respectively, for feature selection. As discussed in~\cite{tei}, taking the difference of channel statistics between two consecutive frames as input for channel-level enhancement learning is more effective for video analysis than the original channel-wise statistics $\{\bm{x}_t\}^{T}_{t=1}$ proposed in Squeeze-and-Excitation Network~\cite{hu2018squeeze}, which is especially designed for image recognition. We choose to use the difference of channel statistics between two consecutive frames as the input of PEM. With the differencing operation, we obtain the difference of channel-wise statistics $\{\bm{d}_t\}^{T}_{t=1}$. The differencing operation is defined as

	\begin{equation}
		\bm{d}_t = f_2(\bm{x}_{t+1}) - f_1(\bm{x}_t) \,, \label{eq:diff}
	\end{equation}
	
\noindent and the difference of the last frame, $\bm{d}_T$, is set as a vector with ones to maintain the magnitude of the memory vector . 

To extend the receptive field of enhancement learning, we here introduce an accumulation operation into the learning of channel-level enhancement for each frame. By the accumulation, the enhancement module of each current frame can be aware of the vital motion information in the previous timings, and not be trapped into the local temporal window as in~\cite{tei}. The accumulated vector $\bm{m}$, named as {\it memory}, accumulates $\bm{d}$ at each time step, and the accumulation operation is controlled by $\gamma \in [0,1]$, as defined in Eq.~(\ref{eq:accumulate}):

	\begin{equation}
		\bm{m}_t = (1-\gamma)\cdot\bm{m}_{t-1} + \gamma\cdot\bm{d}_{t}\,,~~~~~\gamma = \sigma(\bm{W}_g(\bm{m}_{t-1} \mathbin\Vert \bm{d}_{t}))\,.\label{eq:accumulate}
	\end{equation}

\noindent The factor $\gamma$ is determined by the accumulated vector $\bm{m}_{t-1}$ and the difference information $\bm{d}_t$, where $\mathbin\Vert$ denotes a concatenation operation, $\bm{W}_g$ is a projection matrix for linear transformation, and $\sigma(\cdot)$ is a Sigmoid activation function. The final enhancement vector $\bm{a}$ is then generated by

	\begin{equation}
 		\bm{a}_t = \sigma(\bm{W}_a\bm{m}_t)\,, \label{eq:biga}
 	\end{equation}
 
 \noindent where $\bm{W}_a$ is a matrix linearly projecting $\bm{m}$ into a new vector space. With PEM, the network is able to progressively select the motion-related channels in each frame, and adaptively filter the discriminative information for video analysis. The enhanced feature maps $\{\bm{U}_t\}^{T}_{t=1}$ are then fed into a temporal modeling module for temporal information fusion, and we write the output as $\{\bm{Z}_t\}^{T}_{t=1}, \bm{Z} \in \mathbb{R}^{C \times W \times H}$.

\subsection{Temporal Diversity Loss}
\label{sec:tdloss}

It is well-known that feature maps from high-level layers tend to have responses to specific semantic patterns. Convolution kernels that focus on the background of a video may generate similar semantic patterns for the same channels of features from different frames, which may lead to redundant visual feature extraction for video analysis. To calibrate the kernels in 2D CNN and force the network to focus on and capture the variations among frames of a video sequence, we propose the Temporal Diversity Loss to regularize the network toward learning distinguished visual features for different frames. For the feature map $\bm{Z}_t$ from each frame, its $C$ vectorized channel features are denoted as $\{\bm{z}^c_t\}^C_{c=1}$, $\bm{z} \in \mathbb{R}^{WH}$. We use the Cosine Similarity to measure the similarities of a specific channel between two frames of each video frame pair, and then define the loss as:

	\begin{equation}
		\mathcal{L}_\mu = \sum_c \frac{1}{|\mathds{I}|} \sum_{(i,j) \in \mathds{I}} \eta(\bm{z}^c_i,\bm{z}^c_j)\,, \label{eq:tdloss}
	\end{equation}	
	
\noindent where $\mathds{I}=\{(i,j)\,|\,i\neq j,\,1 \leqslant i,j \leqslant T\}$, $\vert\cdot\vert$ indicates the total number of elements in a set, and $\eta(\cdot)$ defines the Cosine Similarity measure, namely $\eta(\bm{x},\bm{y}) = \frac{\bm{x}^\intercal\bm{y}}{{\|\bm{x}\|}_2\cdot{\|\bm{y}\|}_2}$. Considering that the static information among frames is also beneficial to recognition, we only use $C_{\mu}$ ($C_{\mu} < C$) channels for temporal diversity regularization. A further analysis will be discussed in Sec.~\ref{sec:ratio}. With the proposed Temporal Diversity $\mathcal{L}_{\mu}$, the regularization term $\mathcal{L}_r$ is then defined as $\mathcal{L}_r = \sum^{B_{\mu}}_{b=1} \mathcal{L}^b_{\mu}$, where $B_{\mu}$ is the number of blocks with temporal diversity regularization.


\section{Experiments}
\label{sec:experiment}

In this section, the proposed method is evaluated on three benchmark datasets, the Something-Something V1 dataset~\cite{goyal2017something}, Something-Something V2 dataset~\cite{mahdisoltani2018effectiveness}, and the Kinetics400 dataset~\cite{carreira2017quo}. We first briefly introduce these three datasets and the experiment settings in Sec.~\ref{sec:dataset} and Sec.~\ref{sec:setting}, respectively. Then, our method is compared with the state-of-the-arts in Sec.~\ref{sec:sotas}. The ablation study is conducted in Sec.~\ref{sec:ablation} to evaluate the performance of each individual module of our proposed method. In Sec.~\ref{sec:analysis}, we evaluate the proposed method in detail, including parameter, position sensitivity analysis and visualization.

\subsection{Datasets}
\label{sec:dataset}

\subsubsection{Something-Something V1\&V2} \label{sec:something} are crowd-sourced datasets focusing on temporal modeling. In these two datasets, the scenes and objects in each single action category are various, which strongly requires the considered model to focus on the temporal variations among video frames. The V1 \& V2 datasets include 108,499/220,847 videos, respectively, containing 174 action categories in both versions.

\subsubsection{Kinetics400} \label{sec:kinetics} is a large-scale YouTube-like dataset, which contains 400 human action classes, with at least 400 video clips for each category. The average duration of video clips in this dataset is around 10s. Unlike Something-Something datasets, Kinetics is less sensitive to temporal relationships, so the scene information is of importance in its recognition.

\setlength\intextsep{-10pt}
\begin{wrapfigure}{r}{0.5\textwidth}
  \begin{center}
    \includegraphics[width=0.5\textwidth]{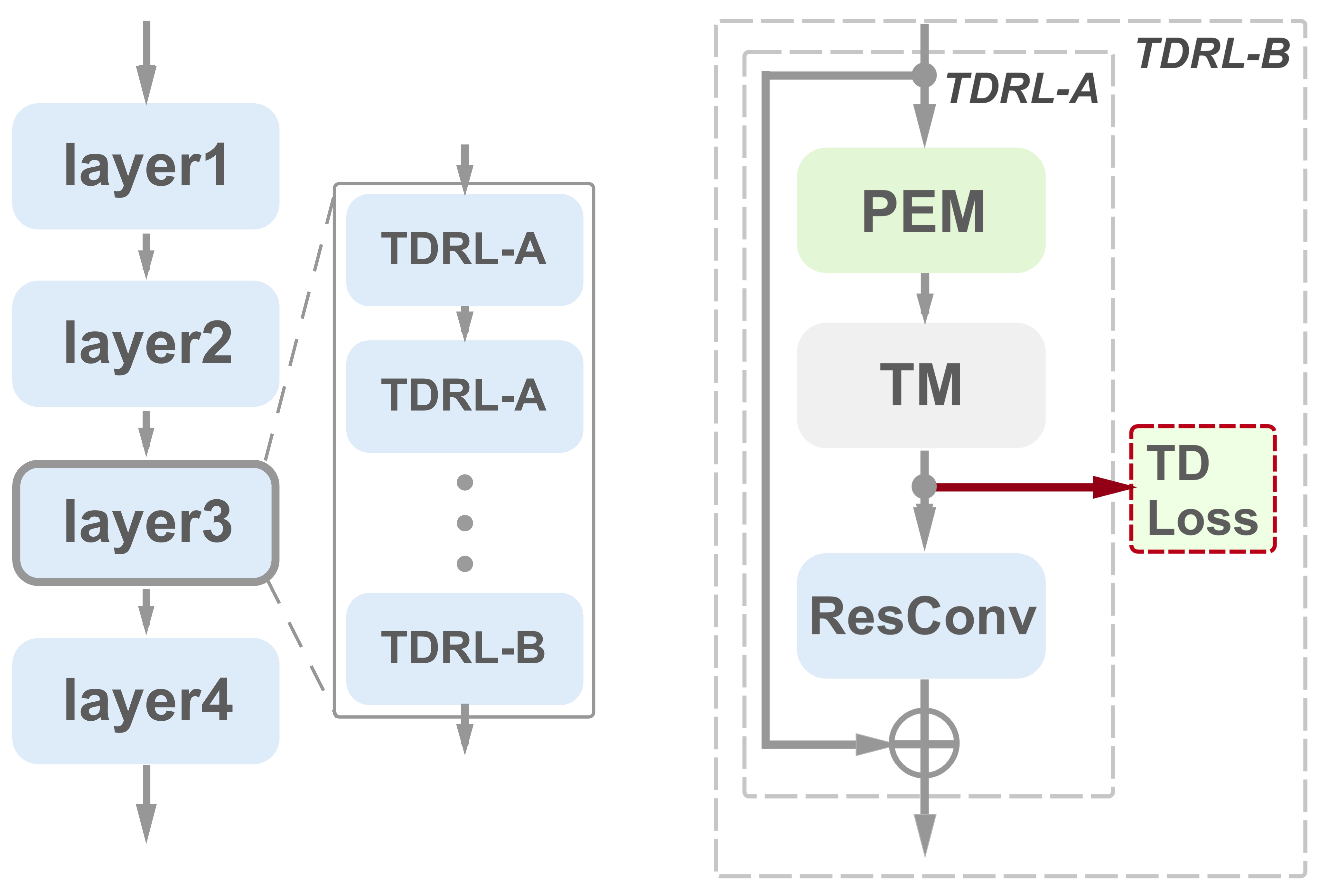}
  \end{center}
  \vspace*{-15pt}
  \caption{Block Illustration}
  \vspace*{12pt}\label{fig:block}
\end{wrapfigure}


%

\subsection{Experimental Setup}
\label{sec:setting}

In all the conducted experiments, we use the ResNet-50~\cite{resnet} as our backbone considering the tradeoff between performance and efficiency, and our model is pre-trained by ImageNet~\cite{imagenet}. We set the Progressive Enhancement Module (PEM) in front of all the blocks of the ResNet backbone. Given that the early stages of the Convolutional Network focus more on spatial appearance modeling and the later ones focus on temporal modeling~\cite{r(2+1)d,gst} and for better convergence, we regularize the temporal diversity of feature maps in the last blocks of each of the last three layers. The Temporal Diversity Loss (TD Loss) is located right after the temporal modeling module. What follows the temporal modeling module is the convolution operation (ResConv) taken from the ResNet block, which includes one $1\times1$, one $3\times3$, and one $1\times1$ 2D convolutions. The position of the PEM and the temporal diversity regularization are illustrated in Fig.~\ref{fig:block}. We define the TDRL-A as the block without the TD Loss, and TDRL-B as the one with the TD Loss, where TDRL stands for Temporal Distinct Representation Learning. The ratio of channels regularized by Temporal Diversity Loss in each feature is 50\%. Without loss of generality, we use the Temporal Interaction Module proposed in~\cite{tei} as our temporal modeling module (TM). $\lambda$ for loss balancing is set as $2\times 10^{-4}$.

\begin{table}[t] 
	\caption{Comparison with the state-of-the-arts on Something-Something V1\&V2 (Top1 Accuracy \%). The notation `I' or `K' in the backbone column indicates that the model is pre-trained with ImageNet or Kinetics400 dataset. The subscripts of `Val' and `Test' indicate the version of the Something-Something dataset. '2S' indicates two streams.}\label{tbl:sth} 
    \begin{center}
      \begin{tabular}{p{2.8cm}<{\centering} p{2.1cm}<{\centering} p{1.4cm}<{\centering} p{1.4cm}<{\centering} p{0.9cm}<{\centering} p{0.9cm}<{\centering} p{0.9cm}<{\centering} p{0.9cm}<{\centering}}
        \toprule 
          \textbf{Method}    & \textbf{Backbone} & \textbf{Frames} & \textbf{FLOPs}  & \textbf{Val$_1$} & \textbf{Test$_1$} & \textbf{Val$_2$} & \textbf{Test$_2$} \\
		\midrule
			I3D~\cite{i3d} & \multirow{3}*{\tabincell{c}{Res3D-50\\(IK)}} & \multirow{3}*{$32f\times2$} & 306G & 41.6 & -- & -- & --\\
			NL I3D~\cite{i3d}    &  & & 334G & 44.4 & --   & -- & --\\
			NL I3D+GCN~\cite{i3d}&  & & 606G & 46.1 & 45.0 & -- & --\\
		\midrule
			\multirow{3}*{TSM~\cite{tsm}} & \multirow{5}*{\tabincell{c}{Res2D-50\\(IK)}} &  $8f$     & 33G & 45.6 & -- & -- & --\\
			& &  $8f\times 2$     & 65G & 47.3 & -- & 61.7 & --\\
			&                             & $16f$     & 65G & 47.2 & 46.0 & -- & --\\
			TSM$_{En}$~\cite{tsm} &                      & $16f+8f$  & 98G & 49.7 & -- & -- & --\\
			TSM-2S~\cite{tsm} &                      & $16f+16f$ & --  & 52.6 & 50.7 & 64.0 & 64.3\\
		\midrule
            \multirow{4}*{TEINet~\cite{tei}} & \multirow{5}*{Res2D-50(I)}  & $8f$ & 33G & 47.4    & -- & 61.3 &  60.6\\
			&& $8f\times 10$ & 990G & 48.8   & -- & 64.0 & 62.7 \\
			& & $16f$ & 66G   & 49.9 & -- & 62.1 & 60.8 \\
			&&  $16f\times 10$ & 1980G & 51.0 & 44.7 & 64.7 & 63.0 \\
			TEINet$_{En}$~\cite{tei}         &                             &  $8f+16f$ & 99G & 52.5 &  46.1 & 66.5 & 64.6\\
        \midrule
        	\multirow{2}*{GST~\cite{gst}} & \multirow{2}*{Res2D-50(I)}  & $8f$ & 29.5G & 47.0    & -- & 61.6 &  60.0\\
			& & $16f$ & 59G   & 48.6 & -- & 62.6 & 61.2 \\
        \midrule
        \midrule
            \multirow{5}*{Ours}  & \multirow{5}*{Res2D-50(I)} & $8f$ &  33G &  \textbf{49.8}  &  \textbf{42.7} & \textbf{62.6} & \textbf{61.4} \\
            &   & $8f\times 2$ &  198G & \textbf{50.4} &  \textbf{--} & \textbf{63.5} & -- \\
            &   & $16f$ & 66G &  \textbf{50.9}  &  \textbf{44.7} & \textbf{63.8} & \textbf{62.5} \\
            &   & $16f\times 2$ & 396G &  \textbf{52.0}  &  \textbf{--} & \textbf{65.0} & -- \\
            &   & $8f+16f$ & 99G &  \textbf{54.3}  &  \textbf{48.3} & \textbf{67.0} & \textbf{65.1} \\
        \bottomrule
      \end{tabular}
    \end{center}
    
  \end{table}

\subsubsection{Pre-processing} We follow a similar pre-processing strategy to that described in~\cite{nonlocal}. To be specific, we first resize the shorter side of RGB images to 256, and center crop a patch followed by scale-jittering. The image patches are then resized to $224\times 224$ before being fed into the network. Owing to the various lengths of video sequences, we adopt different temporal sampling strategies for different datasets. The network takes a clip of a video as input. Each clip consists of 8 or 16 frames. For the Kinetics dataset, we uniformly sample 8 or 16 frames from the consecutive 64 frames randomly sampled in each video. For the Something-Something dataset, due to the limited duration of video samples, we uniformly sample 8 or 16 frames from the whole video.

\begin{table} [t]
	\caption{Comparison with the state-of-the-arts on Kinetics400 (\%). The notations `I', `Z', `S' in the backbone column indicate that the model is pre-trained with ImageNet, trained from scratch, or pre-trained with the Sport1M dataset, respectively. } \label{tbl:k400}
    \begin{center}
      \begin{tabular}{p{3.0cm}<{\centering} p{3.4cm}<{\centering} p{2.8cm}<{\centering} p{1.2cm}<{\centering} p{1.2cm}<{\centering}}
        \toprule 
          \textbf{Method} & \textbf{Backbone}    & \textbf{GFLOPs$\times$views} & \textbf{Top-1} & \textbf{Top-5} \\
        \midrule
            I3D$_{64f}$~\cite{i3d} & Inception V1(I) & 108$\times$N/A & 72.1 & 90.3\\
			I3D$_{64f}$~\cite{i3d} & Inception V1(Z) & 108$\times$N/A & 67.5 & 87.2 \\
			NL+I3D$_{32f}$~\cite{nonlocal} & Res3D-50(I) & 70.5$\times$30 & 74.9 & 91.6 \\
			NL+I3D$_{128f}$~\cite{nonlocal} & Res3D-50(I) & 282$\times$30 & 76.5 & 92.6 \\
			NL+I3D$_{128f}$~\cite{nonlocal} & Res3D-101(I) & 359$\times$30 & 77.7 & 93.3 \\
			Slowfast~\cite{slowfast} & Res3D-50(Z) & 36.1$\times$30 & 75.6 & 92.1 \\
			Slowfast~\cite{slowfast} & Res3D-101(Z) & 106$\times$30 & 77.9 & 93.2 \\
			NL+Slowfast~\cite{slowfast} & Res3D-101(Z) & 234$\times$30 & \textbf{79.8} & \textbf{93.9} \\
			LGD-3D$_{128f}$~\cite{lgdnet} & Res3D-101(I) & N/A$\times$N/A & 79.4 & 94.4 \\
		\midrule
			R(2+1)D$_{32f}$~\cite{r(2+1)d} & Res2D-34(Z) & 152$\times$10 & 72.0 & 90.0 \\
			R(2+1)D$_{32f}$~\cite{r(2+1)d} & Res2D-34(S) & 152$\times$10 & 74.3 & 91.4 \\
			ARTNet$_{16f}$+TSN~\cite{artnet} & Res2D-18(Z) & 23.5$\times$250 & 70.7 & 89.3 \\
			S3D-G$_{64f}$~\cite{s3d} & Inception V1(I) & 71.4$\times$30 & 74.7 & 93.4 \\
			TSM$_{16f}$~\cite{tsm} & Res2D-50(I) & 65$\times$30 & 74.7 & 91.4 \\
			TEINet$_{8f}$~\cite{tei} & Res2D-50(I) & 33$\times$30 & 74.9 & 91.8 \\
			TEINet$_{16f}$~\cite{tei} & Res2D-50(I) & 66$\times$30 & 76.2 & 92.5 \\
		\midrule
			Ours$_{8f}$   & \multirow{2}*{Res2D-50(I)} & $33\times 30$ & \textbf{75.7} & \textbf{92.2} \\
			Ours$_{16f}$  &  & $66\times 30$ & \textbf{76.9} & \textbf{93.0} \\
        \bottomrule
      \end{tabular}
    \end{center}
  \end{table}

\subsubsection{Training} For the Kinetics dataset, we train our model for 100 epochs. The initial learning rate is set as $0.01$, and is scaled with $0.1$ at $50$, $75$, and $90$ epochs. For the Something-Something dataset, the model is trained with 50 epochs in total. The initial learning rate is set as $0.01$ and reduced by a factor of $10$ at $30$, $40$, and $45$ epochs. In the training, Stochastic Gradient Decent (SGD) is utilized with momentum $0.9$ and weight decay of $1\times {10}^{-4}$. The experiments are conducted on {\it Tesla M40} GPUs, and the batch size is set as $64$. The memory vector can be initialized by zeros, or the difference vector $\bm{d}_t$ from the first or last frame, and we experimentally find that the last frame difference $\bm{d}_{T-1}$ can achieve slightly better performance. We therefore use $\bm{d}_{T-1}$ as the memory initialization in the experiments.

\subsubsection{Inference} For fair comparison with the state-of-the-arts, we follow two different data processing settings to evaluate our model. In single-clip ($8$ or $16$ frames) comparison, namely model trained with $8$ frames only ($8f$), or with $16$ frames only ($16f$), we use center cropping for input image processing. The analysis experiments are under the $8f$ setting. In multi-clip comparison, we follow the widely applied settings in~\cite{nonlocal,tsm} to resize the shorter side of images to 256 and take 3 crops (left, middle, right) in each frame. Then we uniformly sample $N$ clips ($8f\times N$ or $16f\times N$) in each video and obtain the classification scores for each clip individually, and the final prediction is based on the average classification score of the $N$ clips.

\subsection{Comparison with State-of-the-Arts}
\label{sec:sotas}

We compare the proposed method with the state-of-the-arts on the Something-SomethingV1\&V2 and the Kinetics400 datasets under different settings for fair comparison. The results are shown in Table~\ref{tbl:sth} and Table~\ref{tbl:k400}, respectively. Table~\ref{tbl:sth} shows that on the Something-Something V1 dataset, our proposed method outperforms the so far best model, TEINet~\cite{tei}, by $2.4\%$, $1.3\%$, and $1.8\%$ under the $8f$, $16f$, and $8f+16f$ settings on the validation set, respectively. Our performance under the two-clips setting is even better than TEINet's performance under the ten-clips setting. On the Something-Something V2 dataset, the performance of our model under the $8f$ setting is even better or the same as the TEINet and GST under the $16f$ setting, which indicates that we can use only half of the inputs of these two models to achieve the same or better accuracy. These results verify the effectiveness of the temporal representation diversity learning. On the Kinetics dataset, the results are reported under the ten-clips-three-crops setting. As can be seen from Table~\ref{tbl:k400}, our proposed model outperforms all the 2D-CNN-based models under different settings, and it even performs better than the 3D-CNN-based nonlocal~\cite{nonlocal} and slowfast~\cite{slowfast} networks with less frames input. We can also witness consistent improvement on the test set of V2, and our model beats TEI by $1.2\%$ under the $8f+16f$ setting.

\begin{table} [t]
    \begin{minipage}[b]{.5\linewidth}
    \caption{Ablation Study - TSM~\cite{tsm} (\%)} \label{tbl:ablationtsm}
    \begin{center}
      \begin{tabular}{l p{1.2cm}<{\centering} p{1.2cm}<{\centering}}
        \toprule
        \textbf{Method}                                       &\textbf{Top-1} &\textbf{Top-5}  \\
        \midrule
            baseline~\cite{tsm} & 45.6 &   74.2   \\
            $+$PEM              & 48.1 &   77.4   \\
            $+$TDLoss           & 47.5 &   76.8   \\
            $+$PEM+TDLoss       & 48.4 &   77.4   \\  
        \bottomrule
      \end{tabular}
    \end{center}
    \end{minipage}
    \begin{minipage}[b]{.5\linewidth}
    \caption{Ablation Study - TIM~\cite{tei} (\%)} \label{tbl:ablationtim}
    \begin{center}
      \begin{tabular}{l p{1.2cm}<{\centering} p{1.2cm}<{\centering}}
        \toprule
        \textbf{Method}                                       &\textbf{Top-1} &\textbf{Top-5}  \\
        \midrule
            baseline~\cite{tei} & 46.1 &   74.7   \\
            $+$MEM~\cite{tei}   & 47.4 &   76.6	\\
            $+$PEM              & 48.7 &   77.8   \\
            $+$TDLoss           & 48.4 &   77.3   \\
            $+$PEM+TDLoss       & 49.8 &   78.1   \\  
        \bottomrule
      \end{tabular}
    \end{center}
    \end{minipage}

  \end{table}

\subsection{Ablation Study}
\label{sec:ablation}

In this section, we evaluate the performances of different modules in our model. We use the single-clip $8f$ setting for the experiment conducted in this section. We use a temporal shift module (TSM)~\cite{tsm} and a temporal interaction module (TIM)~\cite{tei}, respectively, as the baseline in our experiment to show the generality of our model cooperating with different temporal modeling modules. As can be seen from Table~\ref{tbl:ablationtsm}, with the PEM and the TD Loss, there are $2.5\%$ and $1.9\%$ Top-1 accuracy improvements over the TSM, respectively. With both the PEM and TD Loss, the improvement reaches $2.8\%$. Similarly, as shown in Table~\ref{tbl:ablationtim}, PEM gives $2.6\%$ Top-1 accuracy improvement over the baseline, and it outperforms MEM~\cite{tei} by $1.3\%$, which also involves a channel enhancement module. With the TD Loss, there is $2.3\%$ improvement over the baseline. We can see from Table~\ref{tbl:ablationtim} that there is $3.7\%$ improvement over the baseline when both the PEM and TD Loss are applied. One more thing we need to point out is that, after 50 epochs training, the TIM baseline's training accuracy reaches $79.03\%$, while with the PEM and the TD Loss, the training accuracies are down to $77.98\%$ and $74.45\%$, respectively. This training accuracy decline shows that our proposed method can avoid overfitting, and force the model to learn the essential motion cues.

\begin{table} [t]
	\begin{minipage}[b]{.35\linewidth}
    	\caption{TDLoss Ratio (\%)} \label{tbl:ratio}
        \begin{center}
          \begin{tabular}{l p{1.2cm}<{\centering} } 
            \toprule 
              \textbf{Method}                                   &\textbf{Top-1} \\
            \midrule
                baseline  	    & 46.1   \\
                $+25\%$ TDLoss  & 47.7   \\
                $+50\%$ TDLoss  & 48.4   \\
                $+75\%$ TDLoss  & 47.8   \\
                $+100\%$ TDLoss & 47.3   \\
			\bottomrule
          \end{tabular}
        \end{center}
    \end{minipage}
    \begin{minipage}[b]{.32\linewidth}
    \caption{Block Position (\%)} \label{tbl:position}
    \begin{center}
      \begin{tabular}{l p{1.2cm}<{\centering} } 
        \toprule
        \textbf{Order} &\textbf{Top-1} \\
        \midrule
        	TM             &   46.1  \\
        	PEM B. TM      &   48.7  \\
            PEM A. TM      &   49.0  \\
            TDLoss B. TM   &   46.9  \\
            TDLoss A. TM   &   48.4  \\
        \bottomrule
      \end{tabular}
    \end{center}
    \end{minipage}
    \begin{minipage}[b]{.31\linewidth}
    	\caption{Impact of $\lambda$ (\%)} \label{tbl:lambda}
        \begin{center}
          \begin{tabular}{p{1.5cm}<{\centering} p{1.2cm}<{\centering} } 
            \toprule 
              \textbf{$\bm{\lambda}$}                                   &\textbf{Top-1} \\
            \midrule
                $0\times 10^{-4}$  & 46.1   \\
                $1\times 10^{-4}$  & 47.9   \\
                $2\times 10^{-4}$  & 48.4   \\
                $3\times 10^{-4}$  & 47.8   \\
                $4\times 10^{-4}$  & 48.0   \\
			\bottomrule
          \end{tabular}
        \end{center}
    \end{minipage}
    
  \end{table}

\subsection{Detailed Analysis}
\label{sec:analysis}

\subsubsection{Ratio of Channel Regularization}
\label{sec:ratio}

The ratio of channel regularization indicates that how many channels are involved for diversity regularization. We set the ratio from $0\%$ (baseline) to $100\%$ to evaluate the impact of the TD loss in this section. The results are shown in Table~\ref{tbl:ratio}. From this table we can see that when half of the channels are regularized, the model achieves the best performance, $48.4\%$. If all the channels are set under the regularization, the model reaches the lowest accuracy $47.3\%$. This comparison shows that not all the channels require the regularization, and channels without regularization are still useful for the recognition task. However, no matter how many channels are involved in the loss, good improvement is still witnessed over the baseline, TIM.

\subsubsection{Position of Blocks}
\label{sec:position}

In this part, we discuss where to insert the PEM and TD Loss in each block. The two modules are located before (B.) or after (A.) the temporal module (TM) individually to see the impact of position on accuracy. We follow the position of MEM in TEINet~\cite{tei} for fair comparison. The results are shown in Table~\ref{tbl:position}. As can be seen, for PEM, there is no much difference between the two locations. It can fairly provide stable improvement on two different positions. For the TD loss, we discover that when it is located before the TM, the improvement over the baseline is limited. Because TM is inclined to make the representation similar to each other, the following ResConv cannot well extract the video representation. While when the TD regularization is after TM, there is $2.3\%$ improvement over the baseline. The TD loss effectively diversifies the representations among different frames after TM.
\begin{figure}[t]
      \begin{center}
         \includegraphics[width=1.0\linewidth]{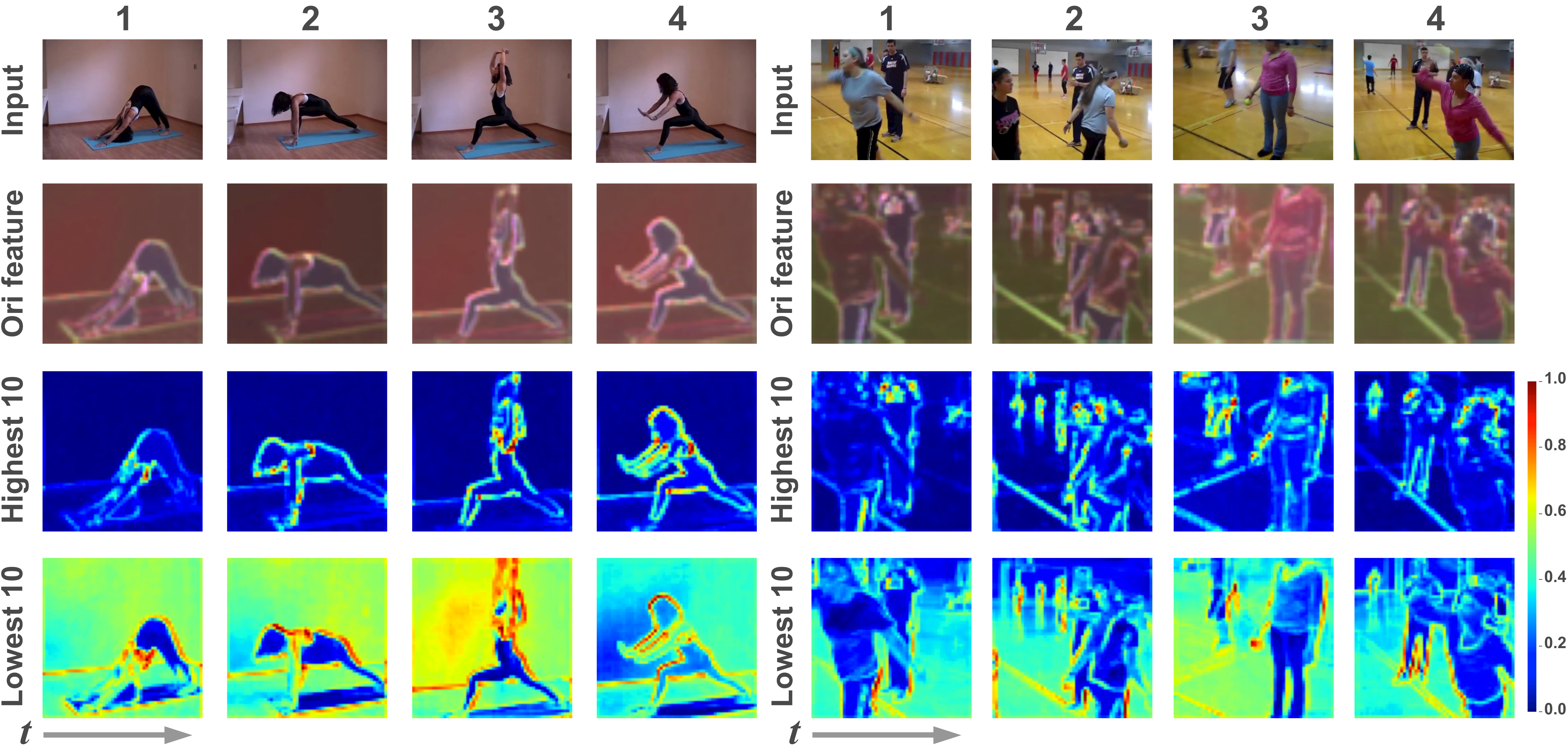}
      \end{center}
         \caption{Visualization of the features excited and suppressed by PEM. The features are organized in the temporal order, and they are from the first block of the first layer. \textbf{R1}: the images in this row are the input images of the network. \textbf{R2}: the features in this row are those before PEM. The channels of each of these features are divided by three groups, and the channels in each group are gathered by average pooling to generate a 3-channels feature, presented as an RGB image. \textbf{R3-4}: Each of the feature map in the these two rows is the average of channels picked from the features before PEM. Each feature map in the third row is gathering of ten channels with the highest enhancement, and each one in the fourth row is gathered with the lowest enhancement.}
      \label{fig:visPEM}
   \end{figure}

\subsubsection{Loss Balancing Factor $\lambda$} 
\label{sec:lambda}

We analyze the impact of the loss balancing factor $\lambda$ on the accuracy in this section. We set $\lambda$ from $1\times 10^{-4}$ to $4\times 10^{-4}$. The result comparisons are shown in Table~\ref{tbl:lambda}. As can be seen, the fluctuation is in the range of $0.6\%$, which shows that the proposed TD Loss is not very sensitive to $\lambda$ when this factor is in an appropriate range. No matter which $\lambda$ we set, the involvement of the TD Loss can still help improve the accuracy over the baseline, which shows the effectiveness of the proposed temporal diversity loss.

\subsubsection{Visualization}
\label{sec:visualization}

\begin{figure}[t]
      \begin{center}
         \includegraphics[width=1.0\linewidth]{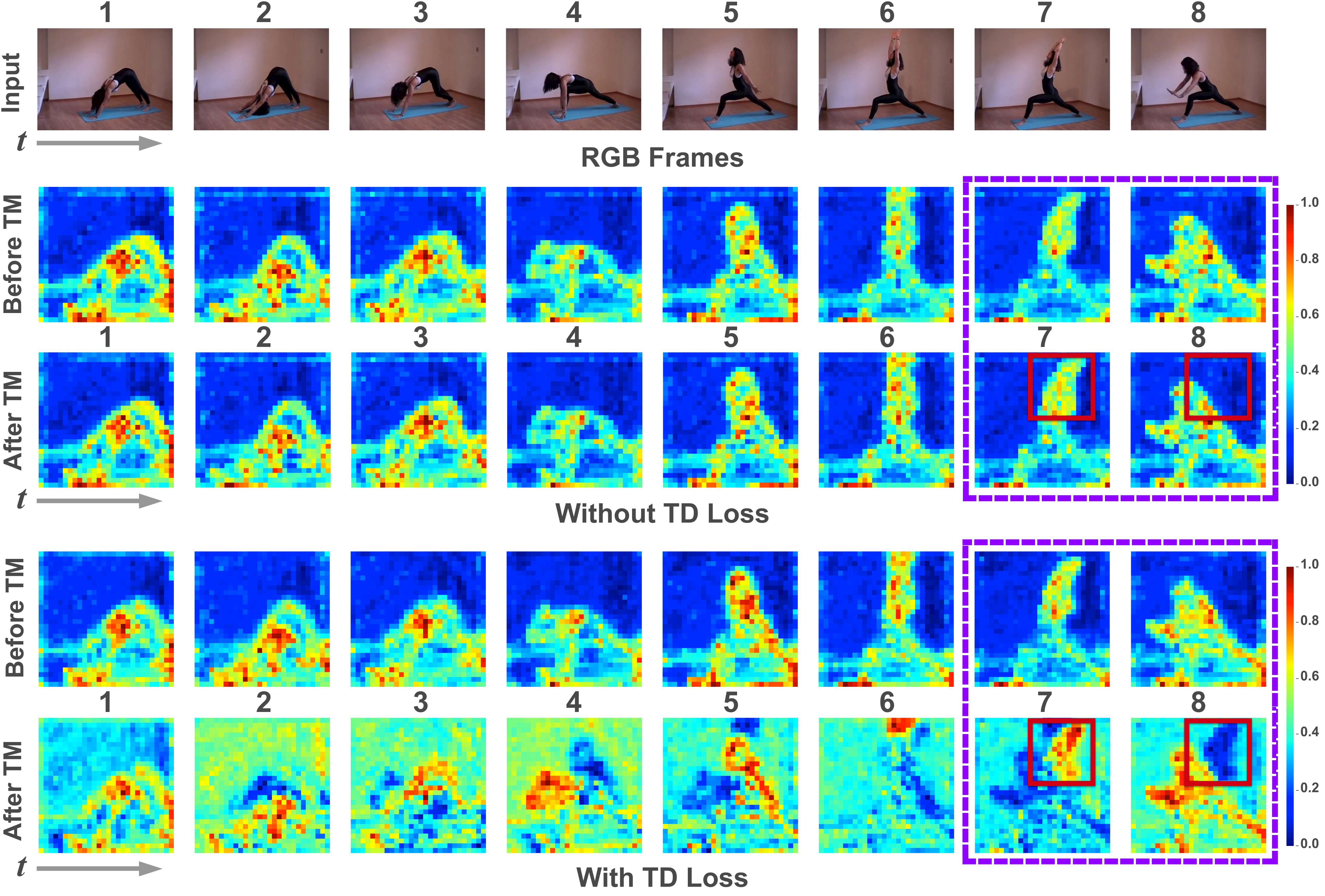}
      \end{center}
         \caption{Visualization of feature maps before and after TM w/ or w/o the TD loss. These feature maps are uniformly sampled from one video and are organized following the temporal order. They are from the last block of the second layer. \textbf{R1}: The images are the input to the network. The purple dashed rectangles mark and illustrate the difference between feature maps with and without TD Loss. \textbf{w/ TD Loss}, the feature maps can well encode action from neighboring frames, and emphasize the variations among them, as marked by red rectangles in the last row. \textbf{w/o TD loss}, the features cannot enhance those variations, as marked by red rectangles in the third row.}
      \label{fig:visTD}
   \end{figure}

We visualize feature maps from different blocks to show the effect of the PEM and TD Loss. The experiment is conducted under the Kinetics dataset. We show the feature maps filtered by the PEM in Fig.~\ref{fig:visPEM}. There are two video samples shown in the figure. The input images are uniformly sampled from a video. From Fig.~\ref{fig:visPEM} we can see that the top ten enhanced channels mainly focus on the motion, while the top ten suppressed channels highly respond to the static background. This visualization shows that the proposed PEM can well discover which are the motion-related channels and which are the repeated static background channels. By enhancing the motion-related ones and suppress the repeated ones, the redundant information can be filtered out and the discriminative one can be well kept. As can be seen from Fig.~\ref{fig:visTD}, with the TD Loss, the feature maps after TM can well encode the information from the current frame and its neighboring frames, while the motion encoded in the features after TM without the TD regularization is very limited. The figures indicate that the TD loss can calibrate the temporal convolution kernels and also enhance the temporal interactions among them.
  
\section{Conclusions}
\label{sec:conclusion}

In this work, we proposed two ways to tackle the issue that the 2D CNN cannot well capture large variations among frames of videos. We first introduced the Progressive Enhancement Module to sequentially excite the discriminative channels of frames. The learned enhancement can be aware of the frame variations in the past time and effectively avoid the redundant feature extraction process. Furthermore, the Temporal Diversity Loss was proposed to diversify the representations after temporal modeling. With this loss, the convolutional kernels are effectively calibrated to capture the variations among frames. The experiments were conducted on three datasets to validate our contributions, showing the effectiveness of the proposed PEM and TD loss.
\subsubsection{Acknowledgement.} We thank Dr. Wei Liu from Tencent AI Lab for his valuable advice.

\clearpage
%
%
\bibliographystyle{splncs04}
\bibliography{egbib}
\end{document}